\documentclass[11pt]{article}

\usepackage[final]{acl}

\usepackage{times}
\usepackage{latexsym}
\usepackage[T1]{fontenc}
\usepackage[utf8]{inputenc}
\usepackage{microtype}
\usepackage{inconsolata}
\usepackage{graphicx}
\usepackage{amsmath}
\usepackage{booktabs}
\usepackage{multirow}
\usepackage{makecell}
\usepackage[inline]{enumitem}
\usepackage{tcolorbox}
\usepackage{algpseudocode} 

\title{Auto-Features: An LLM-Assisted Corpus for Contextual Recognition \\ of Standard and Optional Car Features}

\author{Filippos Ventirozos\textsuperscript{1,2}, Ioanna Nteka\textsuperscript{2}, Tania Nandy\textsuperscript{2}, \\
  \textbf{Peter A. Appleby\textsuperscript{2}, Matthew Shardlow\textsuperscript{1}} \\
  \textsuperscript{1}Manchester Metropolitan University, UK \\
  \textsuperscript{2}Autotrader Research Group, Autotrader UK \\
  \href{mailto:f.ventirozos@mmu.ac.uk,m.shardlow@mmu.ac.uk}{\texttt{\{f.ventirozos,m.shardlow\}@mmu.ac.uk}}}

\begin{document}
\maketitle

\begin{abstract}
We present Auto-Features, a corpus for detecting a car's standard and optional features in advertisement text, given the feature catalogue of that exact vehicle. Sellers name the same feature in many ways. We therefore cast the task as \emph{feature-conditioned}, or contextual, named entity recognition: extract the span that expresses one given catalogue feature. An entity-linking view follows from the same output. The corpus has a human-gold layer (430 adverts) from screened annotators and an LLM-silver layer (925 adverts) auto-annotated by a search-grounded LLM. We also contribute the way it was built. Every LLM output is validated, and each kind of failure has its own remedy. Malformed output is patched. Hallucinated content triggers a fresh attempt by a stronger model. Links to features outside the candidate list are dropped and counted (3 in 10,928 records). The LLM annotates alone only after its agreement with human annotators on link verification is comparable to human--human agreement. On advert-disjoint splits, a LoRA-tuned 4B LLM reaches 63.9\% span F1 and 76.1\% on the entity-linking view. A fine-tuned encoder reaches 59.0\% and 74.1\%. The margin is widest on catalogue features never seen in training (54.5\% vs.\ 49.0\%). 
\end{abstract}

\section{Introduction}

Digital enterprises in the automotive sector increasingly depend on fine-grained information extracted from online car advertisements for market analysis, competitive intelligence, and customer-facing applications. Detecting a car's features informs advert pricing, shows manufacturers which features gain popularity and which tend to fail, and makes listings more transparent for buyers.

Aside from essential components (wheels, doors, drivetrain), cars are equipped with \emph{standard} and \emph{optional} features, and what counts as one is decided per model by each manufacturer: heated seats may be optional on an economy model, standard on a premium model, and absent from the catalogue of a third. There is therefore no global inventory of features to recognise: the label set is specific to the vehicle in each advertisement.

\begin{figure*}
    \centering
    \includegraphics[width=0.66\linewidth]{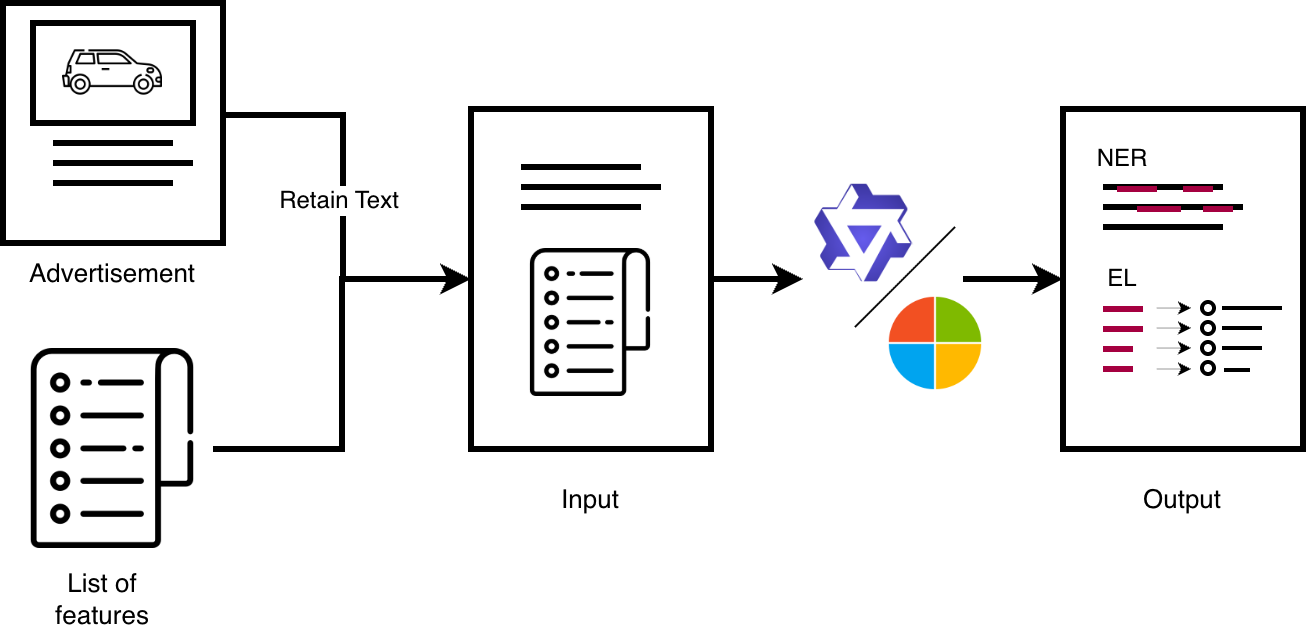}
    \caption{An illustration of our task. The input is the advert text and the list of (optional and standard) catalogue features for that exact vehicle. A model, Qwen3.5-4B LoRA (top) or DeBERTa-v3-base (bottom), labels advert spans with catalogue features (the NER view), from which the set of features present (the EL view) follows.}
    \label{fig:auto-features-illustration}
\end{figure*}

We formalise this as \emph{feature-conditioned} NER. Each advertisement is paired with the catalogue of possible standard and optional features for its exact derivative\footnote{A derivative is the finest classification of a car, following brand, model, and trim level. The catalogue is provided by the manufacturer or configured by domain experts.}, and the task is: given the advertisement text and \emph{one} catalogue feature, extract the token span expressing that feature, or none. We call this \emph{contextual} recognition. The catalogue is supplied with each advert as context, and it decides what counts as an entity. The advertisement's predicted feature set is then the set of catalogue features that received a non-empty span, which yields a binary entity-linking (EL) view of the same predictions at no extra cost. Figure~\ref{fig:auto-features-illustration} illustrates the task. This formulation follows the product attribute--value extraction lineage \citep{zheng2018opentag,yang2022mave}, with the catalogue playing the role of a per-instance, closed-world attribute inventory.

Our central contribution is the corpus, and the way it was built. Wherever a large language model (LLM) takes part, we measure its output against a human reference. (i)~Before the LLM annotates alone, we establish its agreement with human annotators (\S\ref{sec:agreement}), and (ii)~we screen the human annotators against the calibrated LLM and against each other (\S\ref{sec:screening}). (iii)~Validation filters drop hallucinated LLM links, and we count what they drop (\S\ref{sec:silver}). (iv)~Finally, we train on the synthetic labels and test on human-annotated adverts (\S\ref{sec:humantest}). Our contributions are:
\begin{enumerate}
\itemsep0em
    \item \textbf{The Auto-Features corpus}\footnote{\url{https://github.com/FilipposVentirozos/AutoFeatures}}: 1,256 adverts and $\sim$189K (advert, feature) instances per fold. It has two documented layers, a human-gold layer (430 adverts, screened annotators) and an LLM-silver layer (925 adverts, search-grounded auto-annotation). The train/validation/test splits are grouped and advert-disjoint.
    \item \textbf{A measured methodology for LLM-assisted corpus creation}: bootstrap annotation with hallucination recovery, agreement analysis showing the LLM to be as reliable as an additional human annotator, annotator screening, and validated large-scale auto-annotation.
    \item \textbf{A benchmark with long-tail and supervision-quality analysis}: a fine-tuned encoder against a LoRA-tuned small LLM under one metric. Frequency-stratified results isolate generalisation to unseen catalogue features, and we measure what the synthetic supervision is worth against human gold.
\end{enumerate}

The method is not tied to cars. It needs text and a list of candidate labels per text,
as in attribute extraction over a product taxonomy \citep{karamanolakis-etal-2020-txtract}
or NER guided by a list of entity names \citep{fetahu-etal-2022-dynamic}. The same steps then apply: an
LLM first pass, human correction, an agreement check, annotator screening, and
hallucination filtering.

\paragraph{Roadmap.} \S\ref{sec:rel-work} positions the task. \S\ref{sec:corpus} follows
the corpus from annotation to the silver layer and its statistics. \S\ref{sec:eval} benchmarks a
fine-tuned encoder against a small LoRA-tuned LLM. \S\ref{sec:discussion} analyses the errors and draws lessons for
LLM-assisted corpus construction.

\section{Related Work}
\label{sec:rel-work}

\paragraph{NER and attribute--value extraction.}
NER is a foundational sequence labelling task \citep{tjong-kim-sang-de-meulder-2003-introduction}. Our setting adds a \emph{per-instance} reference list that scopes the label space. A mention is an entity only if the catalogue supplied with that instance has a matching entry. \citet{wu-etal-2020-named} feed a global entity dictionary through attention to improve recognition of rare entities, with no instance-specific constraint. Here the per-instance candidate set defines the task.
In DynamicNER \citep{luo-etal-2025-dynamicner} the type list also changes per instance, but every list is derived from one taxonomy of 155 entity types. Ours is a manufacturer's catalogue for one vehicle.
GLiNER \citep{zaratiana-etal-2024-gliner} also reads its types from the input. It is an encoder that scores spans against type names, and it handles types unseen in training.
The question ``does this advert express this attribute, and where?'' also places the task in the product attribute--value extraction lineage. OpenTag \citep{zheng2018opentag} extracts open attribute values from product profiles by sequence tagging. MAVE \citep{yang2022mave} scales attribute-conditioned extraction to a large multi-source product corpus. Auto-Features carries this formulation to a new domain. The attribute inventory there is per-vehicle, and the text is noisy consumer prose rather than curated product titles and descriptions.

\paragraph{LLMs and NER.}
LLMs can also do NER by generating the answer. GPT-NER
\citep{wang2023gptnernamedentityrecognition} prompts an LLM once per entity type. UniversalNER \citep{zhou2024universalner} distils ChatGPT
into smaller models that handle types unseen in
training. Like
feature-conditioned NER, both take their labels from the input at inference time. Unseen
labels matter here too, since most of our test features are rare or unseen in training
(\S\ref{sec:longtail}). As annotators, LLMs match or beat crowd workers on several annotation tasks \citep{gilardi2023chatgpt}. We let the silver-layer LLM annotate alone only after measuring its agreement with human annotators (\S\ref{sec:agreement}).

\paragraph{Automotive NER.}
The automotive domain has received minimal NER attention. The closest is \citet{hu2024named}, who address NER for automotive accessories and parts in Chinese. Our features are manufacturer-defined and need not be physical components (e.g., blind spot assist). ADMit \citep{park2023admit} is an NER model for in-car voice assistants, trained with domain-adversarial and multi-task learning. Its Korean and English data carry 13 classes defined from a car manual, such as convenience function and chassis. FindVehicle \citep{guan2024findvehicle} is a NER dataset of vehicle descriptions from traffic scenes, built for retrieval. Its coarse types (colour, brand, model, location) stop well short of specification detail. \citet{ventirozos2024shifting} annotate English vehicle adverts for history, condition, and sales information. AutoSpecNER \citep{lee-etal-2026-autospecner} annotates vehicle adverts with 15 specification types, such as model, engine and battery capacity. Catalogue features are outside both schemes.

\section{The Auto-Features Corpus}
\label{sec:corpus}

The work is grounded in a collaboration between Manchester Metropolitan University and Autotrader UK under an action research framework \citep{Molineux2018UsingAR}, which provided access to proprietary data and in-house domain expertise.

\subsection{Dataset Origin}
\label{sec:corpus:origin}
We used in-house data from Autotrader UK's large proprietary database of current car advertisements. They are written by \textit{traders} (dealerships with large stock and high turnover) and by \textit{individuals} (typically listing a single car). We sampled traders equally from the highest-volume sellers, which captures template-based and free-form copy. Private listings were sampled in proportion to UK region, so regional variation in British English is represented. For each advertisement, Autotrader UK tracks the car derivative and supplies its complete catalogue of standard and optional features ($\sim$150 per advert on average).

\subsection{Task Definition}
\label{sec:task}
Let $T = \{t_1, \dots, t_n\}$ be the advert's token sequence and $F = \{f_1, \dots, f_m\}$ its feature catalogue. An \emph{instance} is a pair $(T, f_j)$. The output is a tag sequence over $T$ that marks one span expressing $f_j$ (IO scheme), or all-\texttt{O} if the feature is not mentioned. The \textbf{NER view} scores entity-level (span-exact) micro precision, recall, and F1 over instances. The \textbf{EL view} collapses each tag sequence to the binary decision ``is $f_j$ expressed in $T$?'' and scores binary precision, recall, and F1 over instances. Accuracy is uninformative at $\sim$8\% positives. Both views are computed from the same predictions. The EL view is derived, not separately trained. Negative instances follow a closed-world rule: every catalogue feature that no annotator linked is a negative for that advert.

We use IO tags. Each instance covers one feature and holds at most one span, so two spans never touch inside an instance and a \texttt{B} tag would add nothing. In ``heated seats cruise control'', the two features sit side by side, but each is tagged in its own instance. Spans are released as character offsets, so BIO tags over the whole advert can be rebuilt.

\subsection{Human Annotation Campaign}
\label{sec:annotation}
\begin{figure*}[t]
    \centering
    \includegraphics[width=0.68\textwidth]{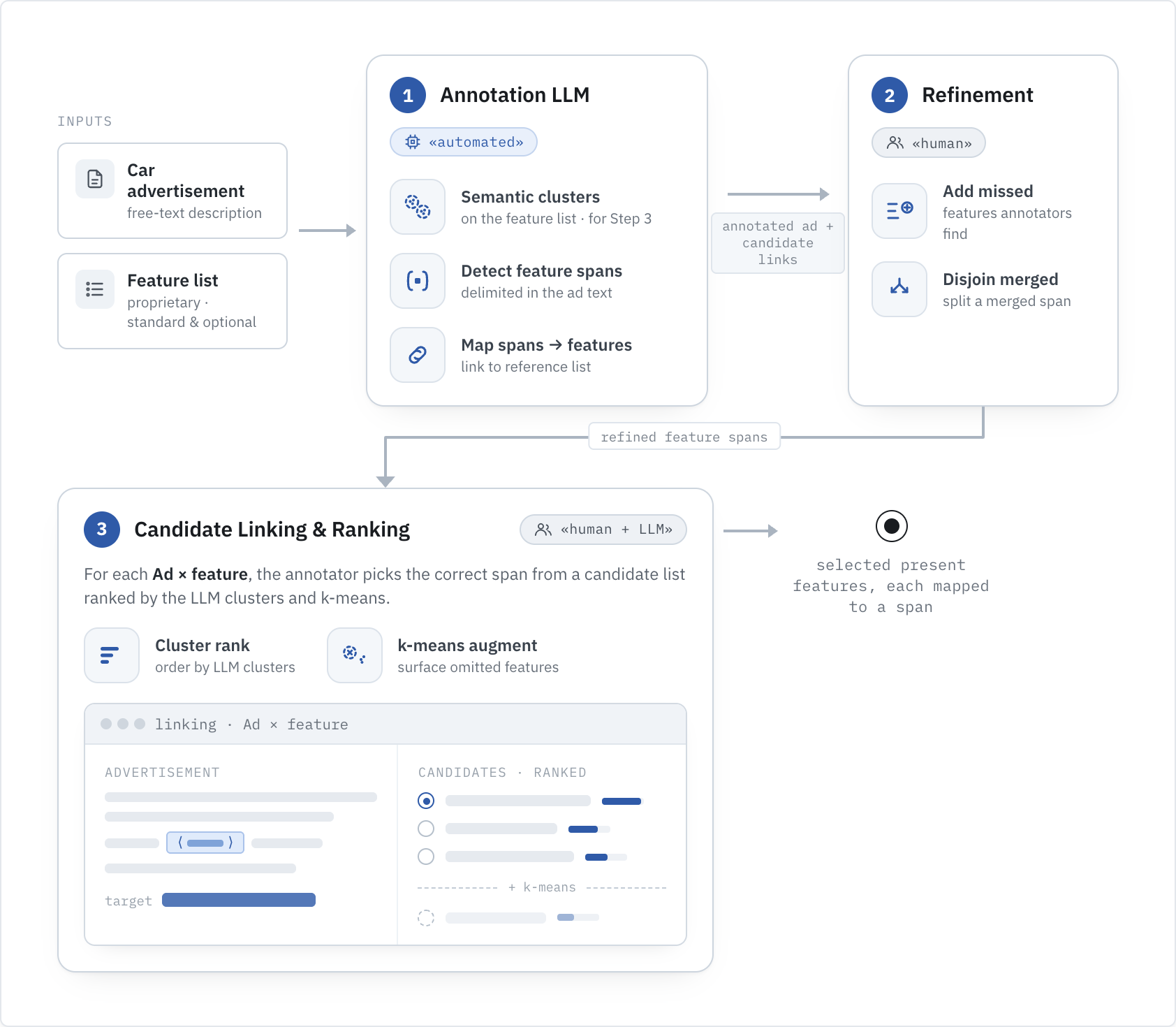}
    \caption{The annotation pipeline. Stage 1 (automated): the annotation LLM clusters the catalogue, detects feature spans in the advert, and proposes candidate links. Stage 2 (human): annotators add missed features and split erroneously merged spans. Stage 3 (human + LLM): given a human-refined mention, annotators select the matching catalogue features from a candidate list ranked by the LLM clusters and k-means, yielding the selected present features, each mapped to a span.}
    \label{fig:annotation_pipeline}
\end{figure*}

Our gold annotation proceeded in three stages (Figure~\ref{fig:annotation_pipeline}): LLM bootstrap annotation, human span refinement, and human link verification.

\paragraph{Stage 1: LLM bootstrap.}
We prompted Gemini models \citep{geminiteam2024geminifamilyhighlycapable} with three instructions. (i)~Mark feature mentions in the text with special characters, similarly to \citet{wang2023gptnernamedentityrecognition}. (ii)~Group the catalogue semantically into clusters. The clusters act as chain-of-thought scaffolding \citep{wei2022chain} and seed the candidate lists shown to annotators (Appendix~\ref{sec:appendix:clustering}). (iii)~Shortlist candidate catalogue features for each detected mention. Output was requested in TOML and validated.\footnote{TOML rather than JSON at this stage: fewer special characters and less escaping to get right, which mattered for the smaller Gemini Flash model driving this step.} A recovery algorithm (Appendix~\ref{sec:appendix:llm}) kept the cost of generation errors low. It started with the smaller Gemini Flash model (\texttt{gemini-2.0-flash-001}) and escalated the prompt, the model (to \texttt{gemini-1.5-pro-002}) or the context only when validation failed. String-level checks removed spurious, duplicated or hallucinated features. Marked spans were mapped back to their original character offsets with a text-alignment library.\footnote{\url{https://github.com/tamuhey/textspan}}

\paragraph{Stage 2: span refinement.}
We recruited over 40 Manchester Metropolitan University students by email to an annotation workshop, run through the university's Rise programme. Most were MSc students and some were PhD students, from a range of STEM programmes, and they earned workshop credits usable towards coursework. None had to be an automotive expert, and all were free to consult web search. Annotators reviewed every bootstrap span in a customised Prodigy interface \citep{prodigy} that shows the vehicle specification beside the text (Figure~\ref{fig:prodigy_ner}, Appendix~\ref{sec:appendix:images}). They marked missed features and split spans that wrongly joined two. In this stage 1,538 adverts were annotated. On the 514 adverts annotated at least three times, three-way inter-annotator agreement on the span-level decisions was 84.7\% micro-F1 (sd 1.6). Pairwise F1 is the appropriate reliability statistic for span annotation, since $\kappa$ needs well-defined negative cases and spans have none \citep{Hripcsak2005}.


\paragraph{Stage 3: link verification.}
For each span confirmed in Stage~2, annotators selected the matching catalogue features from a ranked candidate list of $\sim$16 options. The answer is usually one feature. It can be several, since a catalogue may list the same physical feature under near-duplicate names, or bundle distinct features into one named option pack. The stage produced $\sim$5.4K verification records across 473 adverts (interface in Figure~\ref{fig:prodigy_el}, Appendix~\ref{sec:appendix:images}). We excluded adverts with a high share of unlinked spans, which left \textbf{430 adverts} in the human-gold layer. Within these, 53 of 5,364 Stage-2 spans (1.0\%) never received a Stage-3 link decision. They affect 37 of the 430 adverts. The other 393 are fully verified.

Annotators were briefed separately before each stage and completed a practice round first (Appendix~\ref{sec:appendix:images} gives the full guidance).

\subsection{Establishing the LLM as a Reliable Annotator}
\label{sec:agreement}
To grow past 430 adverts, we needed a calibrated automatic annotator. First we measured whether an LLM was fit for the role.

We re-annotated the Stage-3 linking task with \texttt{gemini-3.6-flash} equipped with Google Search grounding, on the same inputs the humans saw (the identical $\sim$16-option candidate lists), over a curated subset of 1,301 records. Human--human agreement comes from the campaign's 108 doubly-annotated tasks (118 annotator pairs over 40 annotator identities). Because accept-sets are set-valued, we report exact-match, Dice overlap, Cohen's $\kappa$ \citep{cohen1960kappa}, and Krippendorff's $\alpha$ with the MASI distance \citep{krippendorff2019content,passonneau-2006-measuring}, with percentile-bootstrap 95\% confidence intervals. Appendix~\ref{sec:appendix:metrics} defines each metric and what it is sensitive to.

\begin{table}[t]
\centering
\resizebox{\columnwidth}{!}{
\begin{tabular}{lcccc}
\toprule
\textbf{Comparison} & \textbf{Exact} & \textbf{Dice} & \textbf{$\kappa$} & \textbf{$\alpha$ (MASI)} \\
\midrule
Human--human & 52.5\% & 0.686 & \makecell{0.641 \\ {[}0.549, 0.725{]}} & \makecell{0.556 \\ {[}0.463, 0.642{]}} \\
\addlinespace
Human--LLM & 67.2\% & 0.774 & \makecell{0.682 \\ {[}0.653, 0.708{]}} & \makecell{0.679 \\ {[}0.653, 0.706{]}} \\
\bottomrule
\end{tabular}
}
\caption{Agreement on the link-verification task (set-valued accepts; 95\% bootstrap CIs): the LLM agrees with a human annotator at least as well as two humans agree with each other.}
\label{tab:iaa_el}
\end{table}

In Table~\ref{tab:iaa_el}, human--LLM agreement ($\kappa=0.682$) is at least as high as human--human agreement ($\kappa=0.641$). The $\kappa$ intervals overlap. The intervals on $\alpha$, the coefficient designed for set-valued annotation, do not, and human--LLM is the higher. Human--human exact match is only 52.5\%, yet its chance-corrected agreement is \emph{substantial} \citep{landis1977measurement}. Prima facie the LLM looks better, but the two rows rest on different samples. Human--human agreement exists only on the 108 tasks that two humans both did, while the LLM re-annotated all 1,301. We therefore claim only that the LLM is \emph{as reliable as an additional human annotator}.

\begin{table}[t]
\centering
\resizebox{\columnwidth}{!}{
\begin{tabular}{lrr}
\toprule
\textbf{Outcome of a human--human pair} & \textbf{Pairs} & \textbf{\%} \\
\midrule
Identical (non-empty links) & 38 & 32.2 \\
Both linked nothing & 24 & 20.3 \\
Overlapping links, one linked more & 30 & 25.4 \\
Disjoint choices & 26 & 22.0 \\
\bottomrule
\end{tabular}
}
\caption{What human disagreement consists of (118 doubly-annotated pairs).}
\label{tab:disagreement}
\end{table}

Exact match is strict on this task. A record offers about 16 options, and some name the same feature twice. One extra or missing pick is enough to break it. Table~\ref{tab:disagreement} breaks the disagreement down. The \emph{Identical} and \emph{Both linked nothing} rows are full agreement (53\% of pairs). Most of the rest is \emph{under-linking}, where one annotator linked features that the other left unlinked. It covers the \emph{Overlapping links} row and 23 of the 26 \emph{Disjoint choices}.

\begin{figure}[t]
    \centering
    \includegraphics[width=\columnwidth]{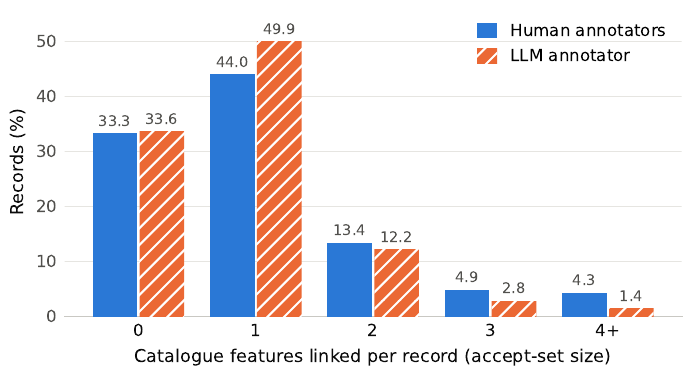}
    \caption{Features linked per record, humans vs.\ LLM (5,440 re-annotated records): near-identical on 0--1 links, with the LLM trimming the many-link tail.}
    \label{fig:accept_dist}
\end{figure}

The same structure shapes the LLM's profile (Figure~\ref{fig:accept_dist}). The figure covers all 5,440 verification records, which the LLM re-annotated for the screening of \S\ref{sec:screening}. On records with zero or one link, humans and the LLM behave almost alike. On the 1,301 calibration records, LLM--human exact agreement on such records is high (82.3\% and 74.2\%), and they cover 78\% of tasks. It drops in the multi-link tail, to 34.1\% at 2--3 accepts and 13.8\% at $\geq$4, where the gold is noisiest. An audit of those disagreements found mostly human bundling of implied features, or outright mislinks.

\subsection{Annotator Quality Screening}
\label{sec:screening}
With the LLM reference established, we turned it on the annotators. The LLM re-annotated all 5,440 verification records. Each of the 44 annotator identities was then scored against the LLM and, where available, against peers. Our pre-registered rule flagged an identity only if it was poor against both references. The peer sample is sparse, so the rule flagged nobody. We therefore removed only the two identities that were clearly poor against the LLM reference alone: a pilot session ($\kappa=0.03$, 6 records) and one annotator with $\kappa=0.23$ (24 records), against a human--human reference of 0.64. Manual review of every affected record confirmed a misreading of the task, well beyond borderline judgement. Both identities linked bundles of unrelated candidates as a habit, e.g., the mention \emph{``Alarm''} to \{\emph{12V Power Socket - Front}, \emph{Steering Wheel - 4 Spoke Leather Trimmed}\}. Their 30 records (across 25 adverts) were replaced by the LLM reference annotation, which preserves coverage. Corpus-level human--LLM agreement rose from $\kappa$ 0.687 to 0.693, and human--human agreement was unchanged. Table~\ref{tab:iaa_el} reports 0.682 on the 1,301-record calibration subset.

\subsection{Silver-Standard Corpus Creation}
\label{sec:silver}
The calibrated LLM annotator (\texttt{gemini-3.6-flash}) then finished what the humans had started. In Stage~2, annotators had refined 9,627 spans that never reached Stage-3 link verification. The LLM linked them, so every silver record combines human span detection with LLM linking. Together with the 1,301 calibration records, the silver layer totals \textbf{10,928 auto-annotated records}. The 9,627 new mentions had no human-prepared candidate lists. We ranked candidates over the advert's full catalogue by embedding similarity (\S\ref{sec:corpus:origin}, $\sim$150 features on average) and shortlisted at most 100 per mention (mean 93). A catalogue with fewer than 100 features is shown in toto.

We developed the prompt and the linking policy on the calibration subset, before any scaling. Appendix~\ref{sec:appendix:prompt} gives the prompt in full, with the optimisation trajectory. The final \emph{evidence-based exhaustive} policy selects all near-duplicate catalogue entries for a mention and resolves vague mentions to the instance this vehicle has. Option packs are linked only on positive evidence. Against the human reference, the policy improved F1 from 0.679 to 0.702 over a precision-first prompt. The web-search grounding was UK-localised. The prompt nudges the model to search whenever it is unsure (Figure~\ref{fig:el_prompt}), yet it did so on only 106 of 10,928 records (1.0\%, 139 queries). De facto, the annotator is the model reasoning over the advert and its catalogue.

Every record passed structural validation, which also dropped hallucinated links (candidates outside the offered list) at write time: \textbf{3 links in 10,928 records} (0.03\%). The resulting linking profile tracks the human population (mean 1.14 accepts per record, Figure~\ref{fig:accept_dist}).

We also ran a preliminary check on the links. A link pairs one mention with one catalogue feature. Two LLMs from other providers, Claude Fable~5.2 and GPT-6 Astra, each judged 1,000 randomly drawn records that carry at least one link, 1,445 links in all. Claude rejected 1.5\% of the links and GPT 2.3\%. Both rejected the same 17 (1.2\%). In all 17 the candidate is not what the advert mentions, as when ``Speed Limiter'' is linked to \emph{Cruise Control}. Missed links were not checked (Appendix~\ref{sec:appendix:judge}).

\subsection{Corpus Statistics and Splits}
\label{sec:data:specs}

\begin{table}[t]
\centering
\resizebox{\columnwidth}{!}{
\begin{tabular}{lrr}
\toprule
\textbf{Layer} & \textbf{Adverts} & \textbf{Instances} \\
\midrule
Human-gold (screened, backfilled) & 430 & 63,481 \\
LLM-silver (grounded auto-annotation) & 925 & 139,680 \\
\midrule
Auto-Features corpus & 1,256 & 188,980 \\
\bottomrule
\end{tabular}
}
\caption{Corpus composition. An instance is one (advert, catalogue feature) pair; $\sim$7.7\% of instances are positive. The two layers overlap on the 99 calibration adverts; where both label the same (advert, feature) pair, one copy is kept.}
\label{tab:stats}
\end{table}

Table~\ref{tab:stats} summarises the corpus. The human-gold layer is rebuilt from the screened annotations (\S\ref{sec:screening}). The silver layer comprises every auto-annotated advert with at least one accepted link.

\paragraph{Grouped splits.}
An advert yields $\sim$150 instances that share its text, so instance-level splitting would place the same advert text in both train and test. We therefore split \emph{by advert}. In each of the three folds, train, validation and test hold disjoint advert sets ($\sim$918/126/212 adverts, with zero overlap verified), balanced on the positive rate.

\begin{table}[t]
\centering
\resizebox{\columnwidth}{!}{
\begin{tabular}{lr}
\toprule
\textbf{Statistic} & \textbf{Value} \\
\midrule
Adverts & 1,256 \\
Words & 190,163 \\
Sub-words (DeBERTa-v3) & 258,759 \\
Words per advert (mean / median / max) & 151 / 123 / 705 \\
\midrule
Distinct spans & 10,062 \\
Words per span (mean) & 3.0 \\
\quad one word (``Bluetooth'') & 9.6\% \\
\quad two words (``Cruise Control'') & 39.7\% \\
\quad three words (``Lane Departure Warning'') & 24.8\% \\
\quad four or more words & 26.0\% \\
\bottomrule
\end{tabular}
}
\caption{Text size and span length. Words are whitespace-separated. Span figures are from fold~1. The span count differs by under 0.2\% in the other folds.}
\label{tab:textstats}
\end{table}

\paragraph{Language.}
All adverts are in British English. The text resembles noisy user-generated corpora such as WNUT \citep{strauss-etal-2016-results} and Twitter NER \citep{ritter-etal-2011-named}: colloquial, telegraphic sales pitches with typos, capitalisation for attention, persuasive language, and dense feature lists. We tokenise with the DeBERTa-v3 SentencePiece tokeniser \citep{he2021debertav3,kudo-richardson-2018-sentencepiece}, whose subword granularity handles the noise and rare tokens. Table~\ref{tab:textstats} gives the text size and the span lengths. Spans average 3.0 words in both layers.

\section{Experiments}
\label{sec:eval}

\subsection{Setup}
\label{sec:exp}
We compare two system classes on folds 1--3 under one metric. Both consume the same initial pre-tokenised text. The LLM sees the space-joined tokens, and its generated span is mapped back to per-token tags. Both are scored by entity-level seqeval F1 on the full-length reference tags, plus the derived EL view of Section~\ref{sec:task}.

\textbf{Encoder.} We fine-tune DeBERTa-v3-base \citep{he2021debertav3} for token classification with the feature prepended: \texttt{[CLS] feature [SEP] advert [SEP]}. The \texttt{token\_type\_ids} mark the feature prefix as context only. No token in it is ever a labelling target, and loss and prediction are restricted to the advert segment. Training runs for at most ten epochs, with early stopping (patience 2) on validation feature-F1. Sequences are capped at 512 tokens.\footnote{2--3\% of test instances exceed the 512-token window; spans beyond it are unreachable for the encoder and count as misses, so both systems are scored against the same references. Scored on window-truncated references instead, the encoder averages 0.4\% absolute higher.}

\textbf{Small LLM.} We tune Qwen3.5-4B \citep{qwen3.5} with LoRA \citep{hu2021loralowrankadaptationlarge} on all linear modules (30.5M trainable parameters, 0.8\% of the model). Decoding is greedy and non-thinking, served by vLLM \citep{kwon2023efficientmemorymanagementlarge}. Training lasts three epochs.\footnote{On fold~1, a schedule stretched over five epochs scored lower (\S\ref{sec:ablations}).} Every epoch's checkpoint is scored by generating on validation, and the best is kept by span F1. We chose a 4B open-weight model for cost and speed (Appendix~\ref{sec:appendix:hyper}). LoRA-tuned open LLMs of this size are already used for NER \citep{zhan-etal-2026-assessment,ewais-etal-2026-just}. We chose LoRA because it comes with published defaults and vLLM serves its adapters.\footnote{vLLM 0.26 does not serve DoRA. Memory was not a constraint, so QLoRA was not needed.} No hyper-parameter was searched. Each is adopted from the literature or measured on this corpus, and Appendix~\ref{sec:appendix:hyper} gives the full configuration, hardware and reproducibility notes.

\subsection{Main Results}
\label{sec:results}

\begin{table*}[t]
\centering
\resizebox{0.92\textwidth}{!}{
\begin{tabular}{l ccc ccc}
\toprule
 & \multicolumn{3}{c}{\textbf{NER view (span F1)}} & \multicolumn{3}{c}{\textbf{EL view (binary, derived)}} \\
\cmidrule(lr){2-4} \cmidrule(lr){5-7}
\textbf{Model} & \textbf{P (\%)} & \textbf{R (\%)} & \textbf{F1 (\%)} & \textbf{P (\%)} & \textbf{R (\%)} & \textbf{F1 (\%)} \\
\midrule
DeBERTa-v3-base (184M, full FT) & 58.1 $\pm$ 1.5 & 59.9 $\pm$ 2.5 & 59.0 $\pm$ 0.7 & 74.1 $\pm$ 3.2 & 74.3 $\pm$ 1.8 & 74.1 $\pm$ 1.0 \\
Qwen3.5-4B + LoRA (30.5M trainable) & \textbf{68.0} $\pm$ 0.7 & \textbf{60.3} $\pm$ 0.6 & \textbf{63.9} $\pm$ 0.6 & \textbf{80.9} $\pm$ 0.5 & 71.9 $\pm$ 0.2 & \textbf{76.1} $\pm$ 0.1 \\
\bottomrule
\end{tabular}
}
\caption{Test results on folds 1--3 (mean $\pm$ sd over folds), identical inputs and metric for both systems; the EL view is derived from the same span predictions. The NER difference is consistent across folds (paired $t(2)=15.9$, $p=0.004$).}
\label{tab:main_results}
\end{table*}

Table~\ref{tab:main_results} gives the main comparison. On the NER view, the LoRA-tuned LLM leads the encoder by 63.9\% to 59.0\% F1, and it leads on every fold (paired $t(2){=}15.9$, $p{=}0.004$, with two degrees of freedom from the three paired folds). On the EL view it leads by 76.1\% to 74.1\%. Precision drives the gap. The LLM reaches 68.0\% against 58.1\%, at similar recall, so it mistakes fewer content words for feature mentions. Rescored with partial credit (exact match 1, overlap 0.5), as in MUC \citep{chinchor-sundheim-1993-muc} and SemEval \citep{segura-bedmar-etal-2013-semeval}, the encoder reaches 63.3\% F1. The two systems sit at different efficiency operating points, and Appendix~\ref{sec:appendix:hyper} reports hardware, throughput and training cost.

\subsection{The Long Tail: Frequency-Stratified F1}
\label{sec:longtail}

\begin{table}[t]
\centering
\resizebox{\columnwidth}{!}{
\begin{tabular}{l r cc}
\toprule
\textbf{Bucket (train freq.)} & \textbf{Support} & \textbf{DeBERTa} & \textbf{Qwen-LoRA} \\
\midrule
frequent ($\geq$50) & 302 & 83.5 $\pm$ 3.0 & \textbf{85.1} $\pm$ 2.3 \\
medium (10--49) & 1,225 & 68.7 $\pm$ 2.2 & \textbf{73.4} $\pm$ 2.8 \\
rare (1--9) & 3,352 & 61.2 $\pm$ 1.4 & \textbf{65.6} $\pm$ 0.9 \\
unseen (0) & 2,804 & 49.0 $\pm$ 1.4 & \textbf{54.5} $\pm$ 0.5 \\
\bottomrule
\end{tabular}
}
\caption{Test F1 (\%) stratified by how often the catalogue feature appears as a positive in that fold's training split (mean $\pm$ sd over folds; support summed over folds).}
\label{tab:stratified}
\end{table}

The catalogue is open-ended, so most test features are barely or never seen in training. \emph{Rare} and \emph{unseen} features make up 80\% of positive test instances, so the aggregate F1 is already a long-tail figure. Table~\ref{tab:stratified} stratifies by training frequency. Both systems degrade monotonically with rarity. The LLM's lead grows down the tail, from +1.6\% absolute on frequent features to +5.5\% on unseen ones. Take the unseen row. Its 2,804 test instances involve catalogue features with no positive training example, yet the LLM still extracts their spans at 54.5\% F1, with a fold-to-fold spread of 0.5\%. A tagger with a closed label set cannot emit a type it never saw. Conditioning on the feature gives both systems a way to do so, and the LLM, with its broader pre-trained knowledge, makes better use of it. A split by \emph{vehicle} instead shows no drop on unseen makes or models (Appendix~\ref{sec:appendix:brand}), so difficulty follows the feature inventory.

\subsection{What Is the Synthetic Supervision Worth?}
\label{sec:humantest}
The main test splits are about 65\% LLM-annotated, so main-fold scores partly measure \emph{agreement with the synthetic annotator}. To separate the two label sources, we built a companion corpus. Training and validation use only the LLM-silver labels (122,772 and 16,908 instances, over 814 and 111 adverts). The test set is 210 human-gold adverts that the silver layer never touched (31,110 instances), and adverts with any LLM-backfilled record are barred from it. The companion corpus is slightly smaller than a main fold, with about 11\% fewer training and 13\% fewer validation instances. Its test set is near-identical in size, with a slightly lower positive rate (7.2\% vs.\ 8.1\%). The numbers can be read side by side. They are not directly comparable, since the test labels come from a different source by design.

\begin{table}[t]
\centering
\resizebox{\columnwidth}{!}{
\begin{tabular}{l ccc c}
\toprule
 & \multicolumn{3}{c}{\textbf{NER view}} & \textbf{EL view} \\
\cmidrule(lr){2-4} \cmidrule(lr){5-5}
 & \textbf{P} & \textbf{R} & \textbf{F1} & \textbf{F1} \\
\midrule
\makecell[l]{Student (Qwen3.5-4B\\trained on the LLM labels)} & 57.3 & 53.3 & 55.2 & 65.2 \\
\makecell[l]{Teacher (the LLM annotator's\\own labels)} & 77.3 & 64.2 & 70.2 & 73.8 \\
\bottomrule
\end{tabular}
}
\caption{Training on synthetic labels only, testing on purely human-annotated adverts (one fold). The teacher row scores the LLM annotator's own labels against the same human gold.}
\label{tab:humantest}
\end{table}

In Table~\ref{tab:humantest}, the \emph{student} is the Qwen model trained on the silver labels, and the \emph{teacher} is the Gemini annotator that produced them. The student reaches 55.2\% F1 on human gold. The teacher scores 70.2\%, but it was given the human-refined mentions and only had to link them. Two checks show how much of the student's score depends on the reference.

The first is the label source. The same student checkpoint scores 68.4\% F1 on validation adverts labelled by the LLM annotator, and 55.2\% on test adverts labelled by humans. Both the adverts and the label source differ, so the 13.2-point gap cannot be attributed to the label source alone. Reweighting the per-bucket F1 by each set's share of frequent, medium, rare and unseen features predicts aggregates only $\sim$1\% apart, so the feature mix explains almost none of the drop. The second is noise in the human reference ($\kappa=0.641$, \S\ref{sec:agreement}). In our audit of student--gold disagreements (\S\ref{sec:discussion}), roughly one third trace to gold under-linking or mislinks, so the measured gap overstates the true one.

The long tail matches Section~\ref{sec:longtail}. Against the full-corpus model of Table~\ref{tab:stratified}, the student holds up on frequent (78.5\%, $-6.6$) and medium (71.1\%, $-2.3$) features. It loses more on rare (57.8\%, $-7.8$) and unseen (43.2\%, $-11.3$) ones.

\subsection{Confirmatory Ablations}
\label{sec:ablations}
We tested two training choices for the Qwen3.5-4B LoRA model on fold~1: how long its learning-rate schedule runs, and where the adapters go. All other settings follow Table~\ref{tab:hyperparams} in Appendix~\ref{sec:appendix:hyper}.

\textbf{LLM epochs.} A second adapter, fine-tuned with the same cosine schedule stretched over 5 epochs instead of 3, scored lower: 63.0\% against 64.5\% on test (Appendix~\ref{sec:appendix:epochs}).

\textbf{LoRA placement.} Qwen3.5 is a hybrid stack, with 24 linear-attention (recurrent) and 8 full-attention layers. The main runs put adapters on all 200 linear modules, the recurrent layers included. Published guidance gives no a priori answer for this architecture. Advice that coverage dominates \citep{dettmers2023qlora} argues for adapting every layer, while hybrid-specific evidence \citep{borobia2026wherelora} argues for freezing the recurrent backbone. We therefore also trained a reduced placement that leaves the linear-attention projections without adapters (128 modules, 21.2M instead of 30.5M trainable parameters). Both adapt the MLPs of all 32 layers. The two tie on test: 64.49\% for the reduced placement and 64.54\% for the full one, inside the $\sim$0.1\% measurement noise floor.\footnote{Batched serving perturbs floating-point summation order, so even greedy decoding is not run-to-run deterministic: three paired replicates of identical weights differed by 0.06--0.09\% F1 on $\sim$19.5K validation rows.} We recommend the reduced placement as the engineering default for span extraction on hybrid stacks, since it trains 30\% fewer parameters. We did not run the remaining folds. They could not resolve a 0.05\% margin.

\section{Discussion}
\label{sec:discussion}

\subsection{Error Analysis}
On the human-gold test set of Section~\ref{sec:humantest}, the student's instance-level disagreements with gold are 962 misses, 674 false alarms and 77 boundary mismatches, out of 31,110 instances. Most errors are about whether a feature is mentioned, not where its span ends. Boundaries still cost some F1, as partial credit shows for the encoder on the main folds (\S\ref{sec:results}). We audited 60 of the disagreements, and the two large classes differ in kind. Many \textbf{false alarms are correct}. The model links \emph{Immobiliser} to the advert token ``Immobiliser'', or \emph{CD Changer} to ``6 Cd Changer'', where the gold has no link. This is the human under-linking seen in Section~\ref{sec:agreement}, and roughly a third of the audited disagreements trace to the gold. \textbf{Misses are mixed.} Real model failures cluster on indirect phrasings (``front and rear sensors'' for \emph{PDC --- Park Distance Control - Rear}) and on long compositional feature names. Other misses are gold mislinks that the model rightly refused, such as a span reading ``DAB Radio'' that an annotator linked to \emph{Exterior Pack - Energy Orange}. Measured F1 thus \emph{understates} true extraction quality. The doubly-annotated subset and the LLM reference layer let corpus users model that noise.

\subsection{Lessons for LLM-Assisted Corpus Construction}
We offer three suggestions to practitioners who build a corpus with an LLM. (1)~\emph{Calibrate the LLM before it works alone.} Have it redo a sample of the human task under the same conditions, with the same mentions and the same candidate lists. Score it with the agreement measures used between humans, and give it a new role only once it matches human--human reliability. The calibration also shows where to trust it least. For us that was the multi-link tail. (2)~\emph{Run structural checks and report what they drop.} We ran two closed-world tests: every accepted link must be among the offered candidates, and every span must align to the text. They caught 3 hallucinated links in 10,928 records, a count corpus users can weigh. (3)~\emph{Measure the supervision as well as the model.} Train on the synthetic labels and test on held-out human gold (\S\ref{sec:humantest}). The result prices the silver layer in F1 against human labels.

\subsection{Use in Other Domains}
\label{sec:domains}
The sine qua non of the task is that each text comes with its own list of candidate labels, as a product listing does with the attributes of its category \citep{karamanolakis-etal-2020-txtract} or a clinical note with the codes on the patient's record. Pair each text with its list, create one instance per label, and the steps of Section~\ref{sec:corpus} apply mutatis mutandis. Catalogue size also matters. Ours hold $\sim$150 features, so the LLM annotator's shortlist of up to 100 covers most of a catalogue, and one instance per label means $\sim$150 instances per advert. With thousands of labels per text, the shortlist would hide most candidates and the instances would multiply, so candidate retrieval would have to carry more of the load. We tested one domain, so transfer remains to be shown.

\section{Conclusion}
\label{sec:concl}
We presented Auto-Features, a two-layer corpus for feature-conditioned NER and entity linking over car advertisements. Every LLM contribution to it was measured before use, through agreement calibration, annotator screening and validated auto-annotation. On advert-disjoint splits, a LoRA-tuned 4B LLM outperforms a fine-tuned DeBERTa-v3-base encoder (63.9\% vs.\ 59.0\% span F1), and the margin is widest on catalogue features never seen in training. A model trained on the synthetic layer alone reaches 55.2\% F1 on purely human-annotated adverts, against 70.2\% for the LLM annotator's own labels. The same task can be posed wherever each text has its own label list (\S\ref{sec:domains}). We hope the corpus serves as a benchmark for long-tail, catalogue-conditioned extraction, and its construction as a template for other LLM-built corpora.

\section*{Limitations}

\paragraph{Reference-label reliability.} As in any human-annotated corpus, the reference labels carry noise at the level of annotator disagreement. Human--human agreement is $\kappa=0.64$, substantial on the usual scale (\S\ref{sec:agreement}). This noise makes our scores conservative (\S\ref{sec:discussion}). In addition, 53 of 5,364 Stage-2 spans (1.0\%) count as absent under the closed-world rule without a Stage-3 decision. Both effects are quantified, and the doubly-annotated subset and the LLM reference layer let corpus users model them directly.

\paragraph{No hyper-parameter search.} All hyper-parameters are adopted from published defaults or measured properties of the corpus (Appendix~\ref{sec:appendix:hyper}). We claim no per-corpus optimality, and we have no learning-rate or LoRA-rank sensitivity results. The encoder was evaluated at the \texttt{base} size only, and a larger encoder may narrow the gap in Table~\ref{tab:main_results}. We compare one encoder and one LLM. Other model families, sizes and parameter-efficient methods were not tested.

\paragraph{Encoder context window.} The encoder cannot reach spans beyond its 512-token window (2--3\% of test instances). We score these as misses so that both systems face the same references, and report the alternative truncated-reference score in Section~\ref{sec:exp}.

\paragraph{Scope.} The corpus is English (UK) and single-domain. It is built on proprietary advertisement text from Autotrader UK. The language findings may not transfer to morphologically rich languages. The LLM annotator (Gemini with search grounding) is a commercial system, and its behaviour may drift.

\section*{Ethics Statement}
We aim for better-informed buyers and better tools for sellers. Reliable feature extraction makes a market with marked information asymmetry more transparent. The corpus derives from advertisements voluntarily published for public viewing. Vehicle specifications are non-personal, and no seller identities are part of the annotation targets. In the released corpus, phone numbers and standard seven-character registration plates are replaced with same-length surrogates. Names, postcodes, websites and email addresses are kept as the sellers published them. Annotators (primarily MSc students with some PhD students across STEM programmes) were recruited by email to an annotation workshop run through the Rise programme, participated with informed consent, were awarded workshop credits, and contributed 2--3 hours per stage with free use of web search.
We used commercial LLMs extensively. Their benefits should be weighed against their energy footprint, so we routed most of the annotation through a small, efficient model and escalated to a larger one only on validation failure.

\section*{Acknowledgements}
We thank Autotrader UK for the data and for funding this work. The annotators took part through Manchester Metropolitan University's Rise programme, and we thank them for their time.

\bibliography{auto_features_used}

\appendix

\section{Feature Clustering for Candidate Lists}
\label{sec:appendix:clustering}
The bootstrap's semantic clusters (Stage 1, instruction ii) seeded the candidate lists shown to annotators: rather than rank a mention against a vehicle's full catalogue from scratch, downstream tooling could shortlist within a small number of semantically related clusters first.

\subsection{Why the raw LLM clusters needed refining}
The raw LLM clusters could not be used as they were. The LLM missed a lot: averaged over the linking records, it left 40\% of a catalogue's features out of every cluster. It also hallucinated. Entries that match no feature in the catalogue came to 7\% of the catalogue's size. A further 1\% of the features were placed in more than one cluster.

\subsection{Refinement procedure}
We refined the LLM clusters with k-means (Euclidean distance) over RoBERTa sentence embeddings \citep{liu2019roberta,reimers-gurevych-2019-sentence}. Each feature the LLM had missed is first attached to the cluster of its most similar clustered feature, by cosine similarity. The target cluster count $k$ is the number of LLM clusters with more than one member, plus one sixth of the number of one-member clusters, a reduction factor chosen for annotation-interface reasons: showing annotators more than a handful of clusters at once clutters the candidate list they see. Centroids are seeded, not randomly initialised: one centroid per multi-member LLM cluster, set to that cluster's mean embedding, plus further centroids drawn at random from the one-member clusters to reach $k$ total. Any singletons left over after k-means converges are grouped into one residual catch-all cluster.

\subsection{Evaluating cluster quality}
We score a clustering with a \emph{cohesion} statistic. For a set $S$ of catalogue features an annotator co-selected on one verified EL record (our ground truth of relatedness), and with $\mathcal{C}_S = \{\mathrm{cluster}(s) : s \in S\}$ denoting the distinct clusters those features fall into,
\[
\text{cohesion}(S) = \frac{|S|}{|\mathcal{C}_S|}\text{,}
\]
i.e.\ the count of co-selected features divided by the number of distinct clusters they fall into. A record whose true features all land in one cluster scores as high as its feature count; one where every feature lands in its own separate cluster scores 1. Seeding k-means with the LLM clusters gives better clusters than the default algorithm. With the same $k$ and embeddings, k-means seeded with the LLM clusters scores 1.27, against 1.15 for the default k-means, whose k-means++ initialisation \citep{kmeans++} draws the starting centroids at random from the data points, favouring points far from those already drawn. The raw LLM clusters score higher still (1.42), even though each accepted feature that they leave unclustered counts as a cluster of its own. They could not be used as they were, because they leave out 40\% of the catalogue.


\section{LLM Bootstrap: Recovery Algorithm}
\label{sec:appendix:llm}
Figure~\ref{algo} shows the hallucination detection and recovery loop of the Stage-1 bootstrap. Full prompts and error-handling procedures ship with the corpus distribution.

\begin{figure*}[h!]
\begin{algorithmic}[1]
\Procedure{ExtractAndRecover}{instance, max\_attempts}
    \State $retry\_count \gets 0$; \; $hallucination\_type \gets$ NONE; \; $last\_output \gets$ null \Comment{no correction needed yet}
    \While{$retry\_count < max\_attempts$}
        \If{$hallucination\_type =$ NONE}
            \State $prompt \gets$ GenerateInitialPrompt(instance.text, instance.features) \Comment{first attempt: plain prompt}
            \State $current\_output \gets$ BaseModel.generate(prompt)
        \Else
            \If{$hallucination\_type =$ TOML\_ERROR}
                \State $prompt \gets$ GenerateTomlFixPrompt(last\_output) \Comment{malformed output: ask for a syntax fix}
            \ElsIf{$hallucination\_type =$ NER\_FORMAT\_ERROR}
                \State $prompt \gets$ GenerateNerFixPrompt(last\_output)
            \ElsIf{$hallucination\_type =$ NER\_CONTENT\_ERROR}
                \State $prompt \gets$ GenerateNerContentFixPrompt(last\_output)
            \ElsIf{$hallucination\_type =$ CONTEXTUAL\_HALLUCINATION}
                \State $prompt \gets$ GenerateInitialPrompt(instance.text, instance.features) \Comment{content hallucination: restart from scratch, not a patch}
            \EndIf
            \State $current\_output \gets$ CorrectionModel.generate(prompt) \Comment{a stronger model handles the fix}
        \EndIf
        \State $last\_output \gets current\_output$
        \State $(validation\_status, parsed\_data) \gets$ ValidateOutput(current\_output, instance.text) \Comment{structural + content checks}
        \If{$validation\_status =$ SUCCESS}
            \State \Return $parsed\_data$
        \Else
            \State $hallucination\_type \gets validation\_status$; \; $retry\_count \gets retry\_count + 1$
        \EndIf
    \EndWhile
    \State \Return FAILURE \Comment{max\_attempts exhausted}
\EndProcedure
\end{algorithmic}
\caption{The iterative hallucination detection and recovery loop of the Stage-1 bootstrap (pseudocode).}
\label{algo}
\end{figure*}

\section{Annotation Interfaces}
\label{sec:appendix:images}
\begin{figure*}[h]
    \centering
    \includegraphics[width=\linewidth]{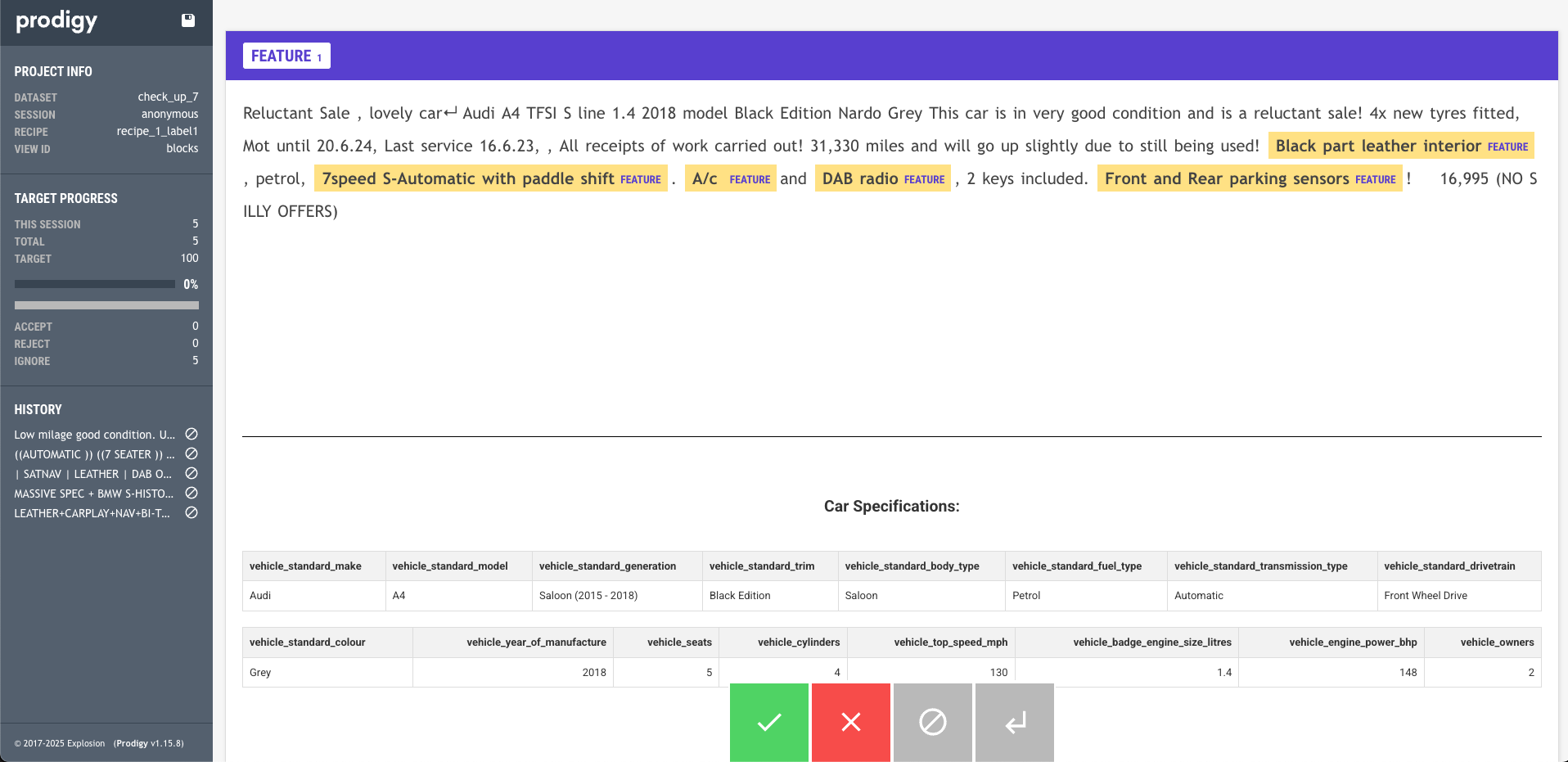}
    \caption{The span-refinement (NER) interface.}
    \label{fig:prodigy_ner}
\end{figure*}

\begin{figure*}[h]
    \centering
    \includegraphics[width=\linewidth]{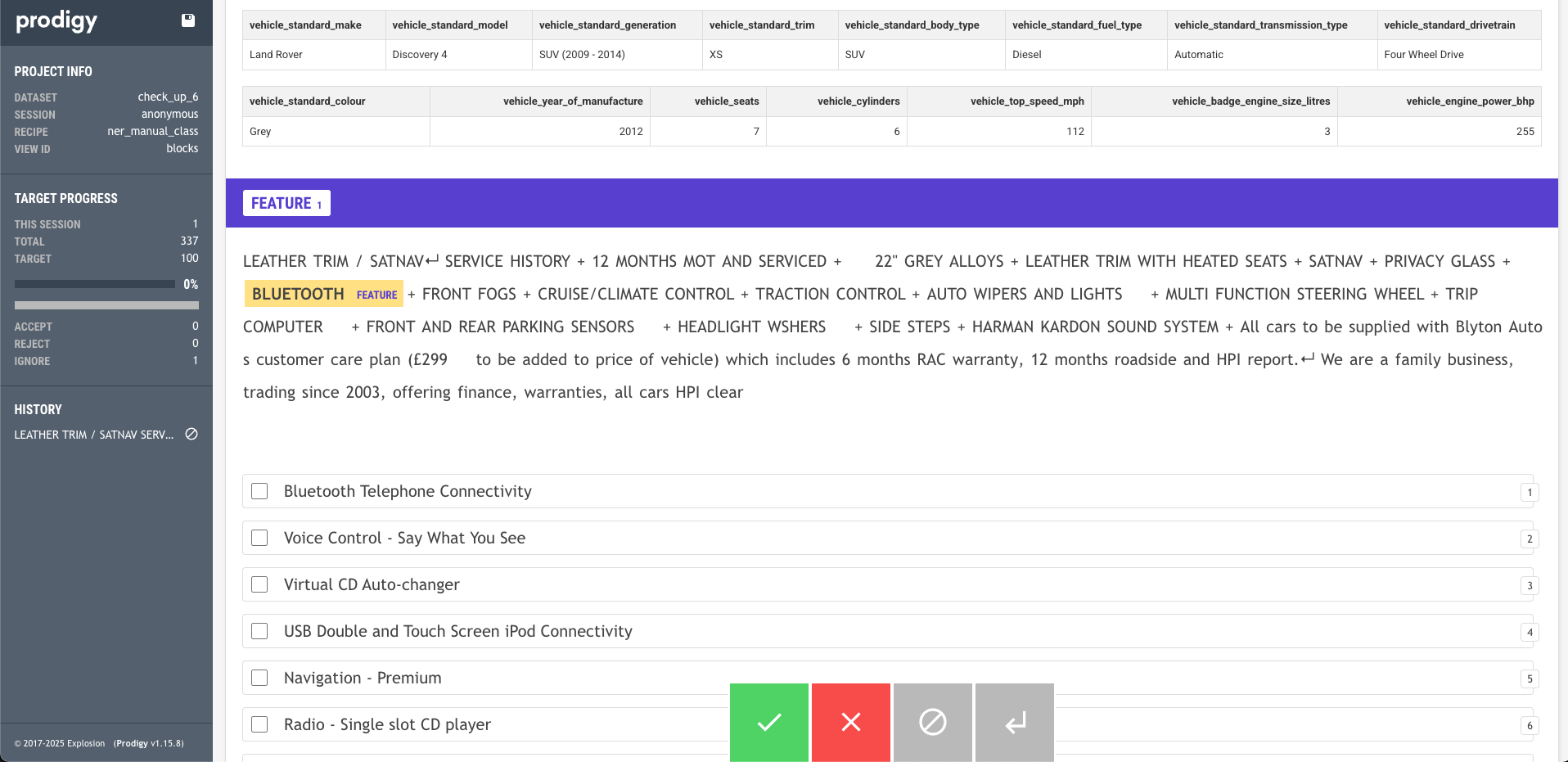}
    \caption{The link-verification (entity linking) interface.}
    \label{fig:prodigy_el}
\end{figure*}

Figures~\ref{fig:prodigy_ner} and \ref{fig:prodigy_el} show the customised Prodigy interfaces for the span-refinement and link-verification stages.

\subsection{Annotator guidance}
Before annotating, every annotator completed a practice round on a small held-out sample, discussed as a group before the real campaign began. For each stage, annotators were briefed separately on what that stage required: Stage 2 annotators were told to mark any missed feature mention and split spans that wrongly joined two features; Stage 3 annotators were told a mention could link to zero, one, or several catalogue features, and were instructed to select every applicable one rather than stop at the first plausible match. This mirrors the exhaustive-but-evidenced policy later encoded in the LLM's own linking prompt (\S\ref{sec:silver}, Figure~\ref{fig:el_prompt}), including its guidance on near-duplicate catalogue entries and option packs, and annotators were briefed on these same quirks. Stage 3 ran roughly a week after Stage 2. Throughout, annotators were free to consult web search and discuss uncertain cases, particularly for less familiar vehicle features.

\section{Agreement Metrics for Set-Valued Annotation}
\label{sec:appendix:metrics}
Each link-verification record gives an annotator's answer as a \emph{set} of accepted catalogue features, possibly empty. All four metrics in Table~\ref{tab:iaa_el} compare the accept-sets $A$ and $B$ of two annotators on the same record. They differ in how much credit partial agreement earns, and in whether agreement by chance is corrected for.

\paragraph{Exact match.} The fraction of records where the two sets are identical, $A = B$. One differing option makes the whole record count as disagreement. It is the strictest and most readable statistic, but it gives no credit for near-misses and no correction for chance.

\paragraph{Dice overlap.} Dice scores each record by $2\,|A \cap B| / (|A| + |B|)$: 1 when the sets are identical, 0 when they share nothing, and a value in between that grows with the overlap. Two empty sets score 1, since both annotators agree the mention has no catalogue match. Dice therefore gives partial credit where exact match gives none, and it equals the F1 of one accept-set scored against the other. Like exact match, it is not chance-corrected.

\paragraph{Cohen's $\kappa$.} Each record is broken into one yes/no decision per candidate: did the annotator accept this option? $\kappa = (p_o - p_e)/(1 - p_e)$ then compares the observed agreement $p_o$ on these decisions against the agreement $p_e$ expected from the annotators' base rates alone \citep{cohen1960kappa}. The correction matters because $\sim$92\% of decisions are ``no''. If one annotator rejects every candidate and the other rejects 92\% of them, the two agree on 92\% of decisions, yet $\kappa$ is exactly 0. Its limitation is that it judges each decision separately and never sees the sets as wholes.

\paragraph{Krippendorff's $\alpha$ with MASI.} $\alpha = 1 - D_o/D_e$ compares observed disagreement $D_o$ against disagreement expected by chance $D_e$ \citep{krippendorff2019content}, so like $\kappa$ it is chance-corrected; unlike $\kappa$, it compares whole sets. The disagreement between two sets is the MASI distance \citep{passonneau-2006-measuring}: their Jaccard overlap $|A \cap B| / |A \cup B|$, discounted further the more the sets depart from a subset relation. Under-linking ($A \subset B$), the dominant human disagreement in Table~\ref{tab:disagreement}, therefore counts as mild, while disjoint choices count as severe.

\section{The Auto-Annotation Prompt and Its Optimisation}
\label{sec:appendix:prompt}

\subsection{Prompt}
Figure~\ref{fig:el_prompt} gives the complete system prompt of the silver-layer annotator, and Figure~\ref{fig:el_user} the user-turn template. The model returns strict JSON (selected candidate numbers and verbatim names, a confidence, and a one-sentence rationale), decoded leniently and validated structurally.

\begin{figure*}[t]
\footnotesize
\begin{verbatim}
You are an expert automotive analyst performing entity linking for used-car classified ads.

You are given:
1. FEATURE MENTION  - a phrase extracted verbatim from a car advert.
2. AD TEXT          - the full advert, for context.
3. VEHICLE SPECS    - the car's specification.
4. CANDIDATE FEATURES - a numbered list of canonical feature names from the manufacturer
   catalogue, ordered most-relevant first.

Task: select EVERY candidate that the FEATURE MENTION refers to. A mention may map to
0, 1, or several candidates. Be exhaustive - prefer including all genuinely matching
candidates over under-selecting.

What counts as a match:
- Paraphrases, synonyms, abbreviations, and spelling/formatting variants of the SAME
  feature (e.g. "hands-free telephone" -> "Bluetooth Interface for Hands-Free Telephone").
- The catalogue often lists the SAME feature under several near-duplicate names; select ALL
  of those duplicates for the mention (e.g. the differently-worded door-mirror entries that
  describe the same mirrors, or two entries that both mean adaptive cruise control).
- Manufacturer-, trim-, or model-specific product/marketing names that denote the
  mentioned feature count as matches.

Being exhaustive without over-reaching:
- A generic or vague mention maps to the specific catalogue instance(s) THIS vehicle
  actually has - use the AD TEXT, SPECS and trim to disambiguate. E.g. "alloys" -> the one
  wheel candidate matching the car's size/trim, NOT every alloy-wheel option in the list.
  Select several instances only when they are genuinely distinct features the car has
  (e.g. "airbags" or "SRS" -> the full airbag set; "heated seats" -> front AND rear if fitted).
- Match at the most specific level the candidate list allows: if a candidate matches the
  mention's specific details (mention "16in alloy wheels" and a 16in wheel candidate exists),
  select that one in preference to a generic candidate. If NO specific candidate fits but a
  generic one covers the mention (the list only has "Alloy Wheels"), select the generic one.
- Only select a PACK or BUNDLE candidate when the AD TEXT or SPECS give positive evidence
  the vehicle actually has that named pack. Do NOT select a pack merely because it could
  contain the mentioned feature.

How to decide:
- Use the AD TEXT, the SPECS and your automotive knowledge first.
- Use the web search tool whenever you are not confident you can correctly interpret a
  manufacturer-, trim-, pack-, or model-specific term - do not guess on those.
- Still do NOT select candidates that are genuinely unrelated or that the mention does not
  denote at all; when truly nothing matches, return an empty list.
- If several mentions are listed, a candidate matches if it denotes ANY of them.

Order selected_numbers and selected_features from most to least relevant.
Return STRICT JSON only, no prose around it.
\end{verbatim}
\caption{The silver-layer annotator's system prompt (final, ``evidence-based exhaustive'' version).}
\label{fig:el_prompt}
\end{figure*}

\begin{figure}[t]
\footnotesize
\begin{verbatim}
FEATURE MENTION: "{mention}"

AD TEXT:
{text}

VEHICLE SPECS:
{specs as "- key: value" bullets}

CANDIDATE FEATURES (choose by number,
0-to-many):
{numbered options}

Return JSON exactly:
{"selected_numbers": [ints],
 "selected_features": [verbatim strings],
 "confidence": float 0-1,
 "reasoning": "one short sentence"}
\end{verbatim}
\caption{The user-turn template. Records carrying several mentions list them all, and a candidate matches if it denotes any of them.}
\label{fig:el_user}
\end{figure}

\subsection{How the prompt was optimised}
The policy was tuned entirely on the 1,301-record calibration subset, scoring each variant against the human reference before any unsupervised annotation:

\begin{enumerate}
\itemsep0em
\item \textbf{Precision-first} (initial): instructed to link only clear matches. P 0.842, R 0.569, F1 0.679: precise but systematically under-linking, replicating the human failure mode.
\item \textbf{Na\"ively exhaustive}: instructed to select every plausible candidate. Recall rose but over-reach appeared: e.g., a bare mention \emph{``alloys''} drew 23 alloy-wheel candidates, and option packs were selected speculatively.
\item \textbf{Evidence-based exhaustive} (final, Figure~\ref{fig:el_prompt}): select every candidate that the advert and the specification support, and no other. Four rules spell this out. (a)~When the catalogue lists the same feature under several names, select them all. (b)~When the mention is vague, as in \emph{``alloys''}, select only the version this vehicle has, read from its specification and trim. (c)~Select an option pack only on positive evidence that the car has it. (d)~Prefer the most specific candidate that fits, and take a generic one only when no specific one does. P 0.787, R 0.633, F1 0.702. On single-link records, the majority of the data, F1 is 0.84.
\end{enumerate}

Two input-engineering steps also measurably helped: the vehicle's HTML specification tables were converted to \texttt{- key: value} bullet lists (4.8$\times$ fewer tokens, 100\% parse rate on the calibration subset), and records whose feature has multiple separate mentions in the advert now list every mention in the prompt rather than just the first occurrence (Figure~\ref{fig:el_user}). Web-search grounding was biased to UK sources to match the market of the adverts.

\section{LLM-as-a-Judge Check of the Silver Links}
\label{sec:appendix:judge}
This check is preliminary. A link pairs one mention with one catalogue feature, and a record can carry several.

\paragraph{Sample.} The silver layer holds 9,627 auto-annotated records that no annotator had linked (\S\ref{sec:silver}). Of these, 7,743 carry at least one link, and we drew 1,000 of them at random with a fixed seed. The draw holds 1,445 links from 471 adverts.

\paragraph{Judges.} The judges were Claude Fable~5.2 and GPT-6 Astra, both at their highest reasoning setting. Web search was off. A judge saw the advert, with personal details masked and the mention marked, the vehicle specification, the full candidate list and the links to assess. The annotation LLM's reasoning and confidence stayed hidden. Both judges answered all 1,000 records.

\paragraph{Criterion.} The judges were told to catch wrong links and to leave the choice of the best link aside. A link is \emph{incorrect} only when the judge is confident that it is wrong. It is \emph{correct} when the candidate is the mentioned feature, it is a variant or pack that the advert does not rule out, or it is a broader feature. The annotation prompt is stricter, since it asks for the variant the car has (Figure~\ref{fig:el_prompt}). We set the criterion on 40 pilot records. They are excluded from the sample.

\paragraph{Results.} Claude did not reject 98.5\% of the links (95\% interval 97.8 to 99.3) and GPT 97.7\% (96.8 to 98.6). Intervals come from a bootstrap over adverts (5,000 resamples). The two verdicts match on 98.6\% of the links.

\paragraph{Why links were rejected.} We read the rationales for the 37 rejected links. In all 17 that both judges rejected, the candidate is not what the advert mentions, as when ``electric mirrors'' is linked to an auto-dimming mirror. Half of the 20 links that only one judge rejected are of the same kind. The rest depend on how strictly the advert is read.

\paragraph{Limits.} Only proposed links were judged, so missed links are out of reach. No human-labelled set was available to calibrate the judges. So the 98\% counts links that were not rejected and should not be read as precision.

\section{Hyper-Parameters, Hardware, and Reproducibility}
\label{sec:appendix:hyper}

\subsection{Hyper-parameters}
Table~\ref{tab:hyperparams} gives the configuration of both systems. No grid search was run.

\begin{table*}[t]
\centering
\resizebox{0.92\textwidth}{!}{
\begin{tabular}{lll}
\toprule
\textbf{Hyper-parameter} & \textbf{Value} & \textbf{Source} \\
\midrule
\multicolumn{3}{c}{\textit{Qwen3.5-4B LoRA}} \\
\midrule
LoRA rank $r$ / $\alpha$ / dropout & 16 / 32 / 0.05 & literature \\
target modules & all linear (200) & tested (\S\ref{sec:ablations}) \\
learning rate / schedule & 2e-4, cosine, warmup 3\% & literature \\
weight decay & 0 & literature \\
epochs & 3, best-epoch on val by generation & tested (\S\ref{sec:ablations}) \\
max sequence length & 2,048 & measured (corpus max 1,641) \\
effective batch (micro $\times$ accum.) & 32 (8 $\times$ 4) & measured \\
precision / decoding & bf16 / greedy, non-thinking & literature \\
\midrule
\multicolumn{3}{c}{\textit{DeBERTa-v3-base}} \\
\midrule
learning rate & 2e-5 & literature \\
epochs & $\leq$10, early stop patience 2 & measured \\
selection metric & val feature-F1 & measured \\
effective batch & 32 & literature \\
max sequence length & 512 & architecture \\
\midrule
seed (both) & 42 & --- \\
\bottomrule
\end{tabular}
}
\caption{Configuration of both systems. Source: \emph{literature} is a published default, \emph{measured} a property of this corpus or hardware, and \emph{tested} an ablated choice.}
\label{tab:hyperparams}
\end{table*}

\subsection{Hardware, software, and efficiency}
Encoder runs: one NVIDIA GB10 (DGX Spark), training all 184M parameters in $\sim$10.7 hours for fold 1, inference at $\sim$59 instances/s. LLM runs: one NVIDIA B200 (183 GB), torch 2.11/cu130, transformers 5.14, peft 0.19, vLLM 0.26, flash-linear-attention 0.5.2; training the 30.5M adapter parameters takes $\sim$3 GPU-hours per fold, generation runs at $\sim$223 instances/s. The two systems ran on different devices, so these throughputs are indicative rather than a controlled comparison; at $\sim$150 instances per advert, the encoder processes about 0.4 adverts per second and the LLM about 1.5, sufficient for batch analytics over listing databases.


\subsection{Fold composition}
Fold 1: train 138,074 instances / 918 adverts (7.8\% positive), validation 19,524 / 126 (6.4\%), test 31,382 / 212 (8.1\%); folds 2--3 analogous. Companion human-test corpus (\S\ref{sec:humantest}): train 122,772 / 814 (LLM labels), validation 16,908 / 111 (LLM labels), test 31,110 / 210 (human labels, 7.2\% positive), drawn from 308 eligible human adverts (98 left unused).

\subsection{LLM epoch probe}
\label{sec:appendix:epochs}
The Qwen3.5-4B LoRA runs train for 3 epochs, with the learning rate decaying from 2e-4 to near zero on a cosine schedule. (The encoder stops early, within 10 epochs, and was not ablated.) On fold~1 we fine-tuned a second adapter with the same schedule stretched over 5 epochs, and scored the checkpoint after every epoch of both runs (Table~\ref{tab:epochs}). The 5-epoch run trails at every epoch (validation F1 52.7 / 49.6 / 59.3 / 63.0 / 62.7 vs.\ 54.6 / 63.5 / 63.7). After epoch 2 it under-predicts badly (precision 72.7\%, recall 37.6\%). Its best checkpoint, after epoch 4, still sits below the 3-epoch endpoint, although it has seen the training data once more. On test it scores 63.0\% against 64.5\%, and the loss concentrates on unseen features ($-2.1\%$), the very generalisation the corpus is built to probe.

\begin{table}[h!]
\centering
\resizebox{\columnwidth}{!}{
\begin{tabular}{l ccccc c}
\toprule
 & \multicolumn{5}{c}{\textbf{Validation F1 (\%) after epoch}} & \textbf{Test} \\
\cmidrule(lr){2-6}
\textbf{Cosine schedule} & \textbf{1} & \textbf{2} & \textbf{3} & \textbf{4} & \textbf{5} & \textbf{F1 (\%)} \\
\midrule
over 3 epochs & 54.6 & 63.5 & \textbf{63.7} & --- & --- & \textbf{64.5} \\
over 5 epochs & 52.7 & 49.6 & 59.3 & \textbf{63.0} & 62.7 & 63.0 \\
\bottomrule
\end{tabular}
}
\caption{LLM epoch probe (Qwen3.5-4B LoRA, fold 1): every epoch's checkpoint evaluated on validation, plus the test score of the checkpoint each schedule selected (bold).}
\label{tab:epochs}
\end{table}

\section{Generalisation Across Vehicle Makes and Models}
\label{sec:appendix:brand}

Does performance depend on having seen the \emph{vehicle}, rather than the feature, in training? We bucket the encoder's test instances by whether their advert's make+model appears among the training adverts of that fold (vehicle metadata accompanies every advert). Table~\ref{tab:brand} shows span F1 for both buckets.

\begin{table}[t]
\centering
\resizebox{\columnwidth}{!}{
\begin{tabular}{l rr rr}
\toprule
 & \multicolumn{2}{c}{\textbf{Model seen in train}} & \multicolumn{2}{c}{\textbf{Model unseen}} \\
\cmidrule(lr){2-3} \cmidrule(lr){4-5}
\textbf{Fold} & \textbf{Adverts} & \textbf{F1 (\%)} & \textbf{Adverts} & \textbf{F1 (\%)} \\
\midrule
1 & 191 & 60.1 & 21 & 55.7 \\
2 & 181 & 58.7 & 31 & 55.3 \\
3 & 170 & 57.9 & 40 & 64.7 \\
\midrule
mean $\pm$ sd & & 58.9 $\pm$ 1.1 & & 58.6 $\pm$ 5.3 \\
\bottomrule
\end{tabular}
}
\caption{Encoder (DeBERTa-v3-base) test span F1 by whether the advert's make+model occurs among that fold's training adverts. Unseen-model support: 185/312/338 gold spans per fold.}
\label{tab:brand}
\end{table}

Unseen vehicles carry no systematic penalty. On average the two buckets score 58.6\% and 58.9\%. The small unseen bucket (21--40 adverts) varies far more from fold to fold than that gap, and in fold 3 it scores highest. At the make level the unseen bucket is too small to interpret (2--4 adverts per fold). This fits the task's design, where difficulty depends on the feature inventory (Table~\ref{tab:stratified}). The corpus can therefore be used to study new-vehicle cold-start without a dedicated split.

\end{document}